\documentclass{elsarticle}

\usepackage{hyperref}

\usepackage{algorithmic}
\usepackage{graphicx}
\usepackage{epstopdf}
\usepackage{epsfig,graphics,subfigure,psfrag,amsmath,amssymb}
\newcommand{\tabincell}[2]{\begin{tabular}{@{}#1@{}}#2\end{tabular}}
\journal{****}

\begin{document}

\begin{frontmatter}

\title{Deep Trans-layer Unsupervised Networks for Representation Learning}

\author[mymainaddress]{Wentao Zhu}
\address[mymainaddress]{Key Lab of Intelligent Information Processing of Chinese Academy of Sciences (CAS), Institute of Computing Technology, CAS, Beijing 100190, China}
\ead{wentao.zhu@vipl.ict.ac.cn}

\author[mymainaddress]{Jun Miao}
\ead{jmiao@ict.ac.cn}

\author[mysecondaryaddress]{Laiyun Qing}
\address[mysecondaryaddress]{School of Computer and Control Engineering, University of Chinese Academy of Sciences, Beijing 100049, China}
\ead{lyqing@ucas.ac.cn}

\author[mymainaddress]{Xilin Chen}
\ead{xlchen@ict.ac.cn}
%
%

\begin{abstract}
Learning features from massive unlabelled data is a vast prevalent topic for high-level tasks in many machine learning applications. The recent great improvements on benchmark data sets achieved by increasingly complex unsupervised learning methods and deep learning models with lots of parameters usually requires many tedious tricks and much expertise to tune. However, filters learned by these complex architectures are quite similar to standard hand-crafted features visually. In this paper, unsupervised learning methods, such as PCA or auto-encoder, are employed as the building block to learn filter banks at each layer. The lower layer responses are transferred to the last layer (trans-layer) to form a more complete representation retaining more information. In addition, some beneficial methods such as local contrast normalization and whitening are added to the proposed deep trans-layer networks to further boost performance. The trans-layer representations are followed by block histograms with binary encoder schema to learn translation and rotation invariant representations, which are utilized to do high-level tasks such as recognition and classification. Compared to traditional deep learning methods, the implemented feature learning method has much less parameters and is validated in several typical experiments, such as digit recognition on \emph{MNIST} and \emph{MNIST} variations, object recognition on Caltech 101 dataset, face verification on LFW dataset. The deep trans-layer unsupervised learning achieves 99.45 \% accuracy on \emph{MNIST} dataset, 67.11 \% accuracy on 15 samples per class and 75.98 \% accuracy on 30 samples per class on Caltech 101 dataset, 87.10 \% on LFW dataset.
\end{abstract}


\end{frontmatter}


\section{Introduction}

Almost all high-layer tasks such as classification, recognition and verification require us to design fine representations for their specific aims. For classification of images taken from the wild, numerous factors in the environment, such as different lighting conditions, occlusions, corruptions and deformations, lead to large amount of intra-class variability in images. Good representations should reduce such non-informative intra-class variability, whilst preserving discriminative information across classes. However, designing good feature representations is a quite tough and difficult procedure for pattern recognition tasks, which is a hot topic in machine learning field.

The research of feature representations mainly contains two aspects, hand-crafted feature designing and automatic feature learning. Researchers and engineers made enormous efforts to devise robust feature representations at their own domains a decade ago. Many successful features are proposed such as SIFT \cite{Lowe_IJCV04} and HoG \cite{Triggs_CVPR05} features in computer vision domain. However, these hand-crafted features have poor transfer ability over domains. Novel features need to be redesigned elaborately when the domain of application is changed. The other way is representation learning, which is a quite prevalent topic after deep learning coming out \cite{Hinton_SCI06}. Nevertheless, these fully learned representations by multi-layer unsupervised learning followed by a fine-tuning procedure have too many parameters to be tuned, and require much expertise knowledge and sophisticated hardware support to train a long time.

In this paper, we demonstrate a novel trans-layer neural network with quite simple and the most classical unsupervised learning method, PCA or auto-encoder, as the building block. Different from the PCANet \cite{Mayi_PCANet}, a one-by-one two layer PCA network, the responses of the previous layer of our model are concatenated to that of the last layer to form a more complete representation. Such trans-layer connections make up the rapid information loss in the cascade unsupervised learning effectively. In addition, the local contrast normalization \cite{LeCun_ICCV09} and whitening are added in our trans-layer unsupervised network to boost its learning ability, which are commonly used in deep neural networks \cite{Coates_KMEANSJ}. The difference between the implemented deep trans-layer unsupervised network and conventional networks is that the deep trans-layer unsupervised network requires no back propagation information to fine-tune the feature banks.

Experimental results indicate that the implemented trans-layer connection scheme boosts the deep trans-layer unsupervised network effectively, and commonly used local contrast normalization and whitening also contribute to the performances. The demonstrated deep trans-layer unsupervised network is validated on digit recognition and object recognition tasks. Quite surprisingly, the stacked conventional unsupervised learning with trans-layer representations achieves 99.45 \% accuracy on \emph{MNIST} dataset, and 67.11 \% accuracy on 15 samples per class and 75.98 \% accuracy on 30 samples per class on Caltech 101 dataset \cite{LiFeifei_CVPR04}.

We will start by reviewing the related works on feature learning and representation in Section \ref{related}. Then the idea of the deep trans-layer unsupervised network, including the pre-processing and trans-layer unsupervised learning, is illustrated detailedly in Section \ref{method}. How to use the deep trans-layer unsupervised network to extract features and tackle applications is also described in Section \ref{method}. The experimental results and comparative analysis on \emph{MNIST}, \emph{MNIST} variations and Caltech 101 datasets are presented on Section \ref{experiment}. Finally, discussions, conclusions and the future work are summarized in Section \ref{discus} and Section \ref{con}.

\section{Related works}\label{related}

Much research has been conducted to pursuit a good representation by manually designing elaborative low-level features, such as LBPH feature \cite{LBPH}, SIFT feature \cite{Lowe_IJCV04} and HoG feature \cite{Triggs_CVPR05} in computer vision field. However, these hand-crafted features cannot be easily adapted to new conditions and tasks, and redesigning them usually requires novel expertise knowledge and tough studies.

Learning good feature representations is probably a promising way to handle the required elaborative and expertise problem in devising hand-crafted features. Much recent work in machine learning has focused on how to learn good feature representations from massive unlabelled data, and great progresses have been made by the methods \cite{Hinton_NIPS12, LeCun_ICML13}. The main idea of deep models is to learn multi-level features at different layers. High-level features generated in the upper-layer are expected to extract more complex and abstract semantics of data, and more invariance to intra-class variability, which is quite useful to high-level tasks. These deep learning methods typically learn multi-level features by greedily ``pre-training'' each layer using the specific unsupervised learning, and then fine-tuning the pre-trained features by stochastic gradient descent (SGD) method with supervised information \cite{Hinton_SCI06}, \cite{LeCun_ICCV09}. However, these deep architectures have numerous parameters such as the number of features to learn and parameters of unsupervised learning in each layer. Besides, the SGD also has various parameters such as momentum, weight decay rate, learning rate, and extra parameters including the Dropout rate or DropConnet rate in recently proposed convolution deep neural networks (ConvNets) \cite{Hinton_NIPS12, LeCun_ICML13}. 

There is also some work on conventional unsupervised learning methods with only single layer \cite{Coates_KMEANSJ, Andrew_NIPS11}. The main idea of these methods is to learn over complete representations with dense features. Although these methods have made much progress on benchmark datasets with almost no hyper parameters, these single layer unsupervised representational learning models require over complete features of dimensions as high as possible, and the parameters need to be elaborately chosen in order to obtain satisfactory results \cite{Coates_KMEANSJ}.

A major drawback of deep learning methods with fine-tuning for stacking representations is the consuming of expensive computational resources and high complexities of the models. One intuition is that, since the elaborately learned features are quite similar to some conventional unsupervised features, such as wavelets, PCA and auto encoder, why not jump over the tough and time-consuming parameter fine-tuning procedure and take those features stacked as the representation directly. Furthermore, more robust invariant features can be better devised other than various pooling strategies. Wavelet scattering networks (ScatNet) are such networks with pre-fixed wavelet filter banks in the deep convolution architectures \cite{Mallat_PAMI13}. The ScatNets have quite solid mathematical analysis of their rotation and translation invariants at any scale. More surprisingly, superior performance over ConvNet and DNNs is obtained by the ScatNets with no fine-tuning phase. However, the ScatNet is shown to have inferior performance in large intra-class variability including great illumination changing and corruption such as face related tasks \cite{Mayi_PCANet}.

The other non-propagation deep network with pre-fixed feature banks is the PCANet \cite{Mayi_PCANet}. The PCANet uses two layer cascaded linear networks with na\"ive PCA filter banks to extract more complex features. The output of the two layer cascaded PCA network is processed by the quantized histogram units. The PCANet presents a superior or highly comparable performance over other methods such as ScatNet \cite{Mallat_PAMI13}, ConvNet \cite{Hinton_NIPS12} and HSC \cite{Yukai_CVPR11}, especially in face recognition tasks with large occlusion, illumination, expression and pose changes. In addition, the PCANet has quite fewer parameters, much faster learning speed and much more reliable than the currently widely researched ConvNets and DNNs, which is much more convenient and more practical to applications \cite{Mayi_PCANet}. However, the cascaded PCA structure in PCANet will face great information loss and corruption after multi-layer transformation, which will be illustrated in Section \ref{discus}. The current prevalent deep networks are also probably facing the same problem that lower layer's discriminative information will be lost after layers' transformation. This leads to inferior results of PCANet on conventional object recognition tasks.

This paper will tackle the multi-layer information loss problem by demonstrating a novel trans-layer structure based on multi-layer conventional unsupervised filter banks. The local contrast normalization and whitening operations are applied to ameliorate the unsupervised learning in the deep trans-layer network. Thus the trans-layer unsupervised network forms a more complete and effective representation, whilst retaining the advantages such as fewer parameters, faster learning speed and more reliable performance. Also, histogram operation is adopted to preserve translation and rotation invariance after binary quantization. Experimental results confirm that the deep trans-layer unsupervised network boosts the performance of conventional unsupervised learning, and it learns effective feature representations that achieve 99.45 \% accuracy on \emph{MNIST} dataset, and 67.11 \% accuracy on 15 samples per class and 75.98 \% accuracy on 30 samples per class on Caltech 101 dataset.

\section{Deep trans-layer unsupervised network}\label{method}

\begin{figure}[!t]
\centering
\includegraphics{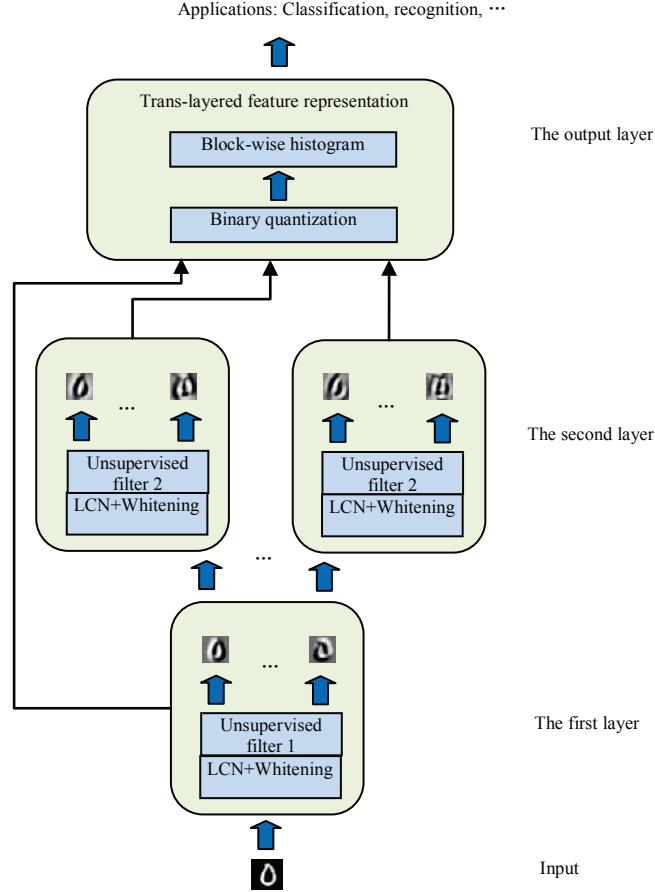}
\caption{The framework of the deep trans-layer unsupervised network, which is a three-layer neural network, including the first unsupervised learning layer, the second unsupervised learning layer and the trans-layer.}
\label{DeepPCAFram}
\end{figure}

In this section, we present a novel framework, the deep trans-layer unsupervised network, for feature learning and representation. The framework of proposed deep trans-layer unsupervised network is illustrated in Figure \ref{DeepPCAFram}. The procedures of the deep trans-layer unsupervised network is similar to other commonly used frameworks in computer vision \cite{LeCun_CVPR10} and feature learning work \cite{Andrew_ICML09} as well. Different from the traditional methods, the deep trans-layer unsupervised network utilizes the unsupervised learning methods, such as the PCA or auto encoder, to learn the local receptive filter banks, and needs no fine-tuning procedure to adjust those local filter banks. Besides, the previous layer's unsupervised feature maps are concatenated to the last layer to form a much more completed representation, which is shown quite effective for the following tasks.

As a high level, the proposed deep trans-layer unsupervised performs the following steps to learn a feature representation.

\textbf{The first layer}

\begin{itemize}
\item Extract random patches from the training images.
\item Apply local contrast normalization (LCN) and whitening operations to each extracted patches.
\item Learn the first layer's filter banks by the unsupervised learning, such as PCA or auto encoder, on the extracted patches.
\item Use the learned feature banks to generate the feature maps of each training image.
\end{itemize}
\textbf{The second layer}
\begin{itemize}
\item Extract random patches from the obtained feature maps in the first layer.
\item Apply local contrast normalization and whitening operations to each extracted patches from feature maps.
\item Learn the second layer's filter banks by the unsupervised learning on the extracted feature map patches.
\end{itemize}
Given test images, we can extract stacked and concatenated feature representations and apply them to classification.

\textbf{The first layer}
\begin{itemize}
\item Apply the unsupervised learning feature banks the first layer learned to each test image to generate its feature maps after local contrast normalization and whitening processing operations.
\end{itemize}
\textbf{The second layer}
\begin{itemize}
\item Apply the unsupervised learning feature banks the second layer learned to each first layer feature maps after local contrast normalization and whitening processing operations.
\end{itemize}
\textbf{The output layer}
\begin{itemize}
\item Generate the response maps by amalgamating the first layer feature maps with the second layer feature maps.
\item Use binary coding operation to facilitate the next histogram procedure.
\item Compose block-wise histogram of the coded response maps of each test image as its feature representation.
\item Apply dimensionality reduction methods or distance metrics to the learned deep trans-layer unsupervised feature representations, or directly train a classifier based the representations to tackle the applications.
\end{itemize}
In the following paragraphs, we will describe the components of the above network pipeline in more details.

\subsection{Representation learning}

The structure of deep trans-layer unsupervised network is partially similar to convolution neural network where convolution operations are done in small patches. In deep trans-layer unsupervised network, the local filters are learned by unsupervised learning, such as PCA or auto encoder, which requires no fine tuning process through error back propagation.

In each unsupervised layer (the first and second layer), the system begins with extracting a large number of random patches from unlabelled input images. Suppose the images used here are all gray images. Each patch has a receptive field size or dimension of \(k_1\)-by-\(k_2\). If the images are color images with \(d\) channels, the patch dimension is \(k_1\)-by-\(k_2\)-by-\(d\). Just process the other \(d-1\) channels the same as following procedures step by step independently. Then a dataset of \(m\) patches is constructed, \(\textbf{X}=\{ \textbf{x}^{(1)},\dots,\textbf{x}^{(m)} \}\), where \(\textbf{x}^{(i)} \in R^{k_1 \times k_2}\) stands for the \(i\)th patch extracted from the input images in the first layer or the feature maps in the second layer. Sequentially, we apply the preprocessing of local contrast normalization (LCN) and whitening, and then unsupervised learning in the first and second layer, respectively.

\subsubsection{Local contrast normalization and whitening}\label{sec:preprocess}

In the pre-processing of each layer's unsupervised learning, we perform several simple operations effectively in the implemented network.

The first is local contrast normalization (LCN) \cite{LeCun_ICCV09}. For each local patch \(\textbf{x}^{(i)}\) in the extracted patch dataset \(\textbf{X}\), we normalize the patch \(\textbf{x}^{(i)}\) by subtracting its mean and dividing by its standard deviation as,
\begin{equation}
\begin{array}{l}
 {\textbf{y}^{(i)}}_{j,k} = ({\textbf{x}^{(i)}}_{j,k} - \frac{1}{{{k_1}{k_2}}}\sum\limits_{j = 1}^{{k_1}} {\sum\limits_{k = 1}^{{k_2}} {{\textbf{x}^{(i)}}_{j,k}} } )/ \\
 (\sqrt {\frac{1}{{{k_1}{k_2}}}\sum\limits_{j = 1}^{{k_1}} {\sum\limits_{k = 1}^{{k_2}} {({\textbf{x}^{(i)}}_{j,k} - \frac{1}{{{k_1}{k_2}}}\sum\limits_{j = 1}^{{k_1}} {\sum\limits_{k = 1}^{{k_2}} {{\textbf{x}^{(i)}}_{j,k}} } } } {)^2}}  + C), \\
 j = 1, \cdots , {k_1}; k = 1, \cdots , {k_2}; i = 1, \cdots , m, \\
 \end{array}
\end{equation}
where \(C\) is a constant integer to make the model more robust, which is commonly manipulated in practice. In our work, \(C\) is set to 10.

The LCN has explicit explanations both in physics and physiology. The mean of local patch stands for local brightness, and the standard deviation represents contrast normalization. By LCN, the illumination and material optimal property effects are removed. On the other hand, the LCN has an effect similar to lateral inhibition found in real neurons. The LCN operator inhabits the responses within each local patch, whilst activating responses in the same location of these patches.

Following the LCN, whitening is the second preprocessing method for each unsupervised learning layer. Whitening is commonly used in various application and is a decorrelation operator, which reduces redundant representation of images. The whitening operator transforms the patches as,
\begin{equation}
\begin{array}{l}
\left[ {\textbf{D}, \textbf{U}} \right] = eig(cov(\textbf{Y})) \\
{\textbf{z}^{(i)}} = \textbf{U}(\textbf{D} + diag(\epsilon)) ^ {-1/2} \textbf{U}^{T}\textbf{y}^{(i)}, i = 1, \dots ,m, \\
\end{array}
\end{equation}
where \(\textbf{Y}\) is formed by \(m\) patches \(\textbf{y}^{(i)}\), \(cov()\) stands for covariance function and the size of output data is \(k_1 * k_2\), \(eig()\) is the eigenvalue decomposition function, \(\textbf{D}\) and \(\textbf{U}\) are eigenvalues and eigenvectors respectively, \(\epsilon\) is set as \(0.1\) here to make the operator more robust. The ZCA whitening also has biological explanation and has been proved its effectiveness by a lot of work.

\subsubsection{Trans-layer unsupervised learning}

After pre-processing for each layer, unsupervised learning, such as PCA or auto encoder, is used to learn feature banks in the trans-layer network.

\textbf{The first unsupervised layer}

Assuming that the number of feature banks in the first layer is \(L_1\), flatten each pre-processed patch \(\textbf{z}_{1}^{(i)}\) extracted from input images, and put the flattened vectors together. Extracted patch matrix from the first layer will be obtained as
\begin{equation}
{\textbf{Z}_1} = \left[ {\begin{array}{*{20}{c}}
   {{\textbf{z}_1}^{(1)},} & {{\textbf{z}_1}^{(2)},} & { \cdots ,} & {{\textbf{z}_1}^{(m)}}  \\
\end{array}} \right] \in {R^{{k_1}{k_2} \times m}}.
\end{equation}

If the used unsupervised learning is PCA, the PCA unsupervised learning aims to
\begin{equation}\label{pcamodel}
\begin{array}{l}
 \mathop {\min }\limits_{{\textbf{W}_1} \in {R^{{L_1} \times {k_1}{k_2}}}} {\left\| {{\textbf{Z}_1} - {\textbf{W}_1}^T{\textbf{W}_1}{\textbf{Z}_1}} \right\|^2}_{{2}} \\
 \begin{array}{*{20}{c}}
   {s.t.} & {{\textbf{W}_1}{\textbf{W}_1}^T = {\textbf{I}_{{L_1}}}}  \\
\end{array}, \\
 \end{array}
\end{equation}
where \(\textbf{W}_1\) is the PCA transformation weights, and \(\textbf{I}_{L_1}\) is an identity matrix of size \(L_1 \times L_1\). We get the solution of the above PCA constraints as
\begin{equation} \label{pcasolution}
{\textbf{W}_1} = {\left[ {\begin{array}{*{20}{c}}
   {{\textbf{w}_1}^1,} & {{\textbf{w}_1}^2,} & { \cdots ,} & {{\textbf{w}_1}^{{L_1}}}  \\
\end{array}} \right]^T},
\end{equation}
where
\(\begin{array}{*{20}{c}}
   {{\textbf{w}_1}^1,} & {{\textbf{w}_1}^2,} & { \cdots ,} & {{\textbf{w}_1}^{{L_1}}}  \\
\end{array}\) are the first \(L_1\) eigenvectors of \(\textbf{Z}_1 \textbf{Z}_1^T\) with \(L_1\) largest variances. Then, we should make these eigenvectors changing back to \(k_1 \times k_2\) dimensions to facilitate the next PCA feature mapping operation.

The first PCA unsupervised layer's feature maps are generated by applying these \(L_1\) filters to convolute with the input images. For each input image, \(L_1\) feature maps can be obtained by
\begin{equation} \label{pcamapping}
\begin{array}{*{20}{c}}
   {{{\bf{I}}_1}^{(i)} = {\bf{I}} * {{\bf{w}}_1}^i,} & {i = 1, \cdots ,{L_1}}  \\
\end{array},
\end{equation}
where \(\bf{I}\) stands for an input image with zero padded to make \({\bf{I}}_1^{(i)}\) have the same size as the input image, and \(*\) stands for a convolution operator.

If the used unsupervised learning is auto encoder, the auto encoder unsupervised learning aims to
\begin{equation}\label{encodermodel}
\begin{array}{l}
 \mathop {\min }\limits_{{\textbf{W}_1} \in {R^{{L_1} \times {k_1}{k_2}}}, \textbf{b}_1 \in {R^{L_1}}} C {\left\| {{\textbf{Z}_1} - \sigma \left( {\textbf{W}_1}^T \sigma \left( {\textbf{W}_1}{{\bf{\tilde{Z}}}_1} + {\bf{b}}_1 \textbf{i} \right) + {\bf{b}'}_1 \textbf{i} \right)}  \right\|^2}_{{2}} + {\left\| {{\bf{W}}_1} \right\|^2}_{{2}}
 \end{array}
\end{equation}
where \(\textbf{W}_1\) is the auto encoder transformation weights, \({\bf{b}}_1\) is the encoder bias, \(C\) is the tradeoff between errors and model complexity, the used activation function \(\sigma()\) is the hyperbolic tangent function, \({\bf{b}'}_1\) is the decoder bias, \(\bf{i}\) is a column vector of size \(m\) full of elements \(1\),  and \({\bf{\tilde{Z}}}_1\) is obtained by randomly turning \(10 \%\) of the elements in \({\bf{Z}}_1\) into \(0\). Then the encoder weights \({\bf{W}}_1\) and bias \({\bf{b}}_1\) are calculated by stochastic gradient decent method. The solution of the above de-noising auto encoders is as
\begin{equation}\label{encodersolution}
{\textbf{W}_1} = {\left[ {\begin{array}{*{20}{c}}
   {{\textbf{w}_1}^1,} & {{\textbf{w}_1}^2,} & { \cdots ,} & {{\textbf{w}_1}^{{L_1}}}  \\
\end{array}} \right]^T}, \\
{\textbf{b}_1} = {\left[ {\begin{array}{*{20}{c}}
   {{b_1}^1,} & {{b_1}^2,} & { \cdots ,} & {{b_1}^{{L_1}}}  \\
\end{array}} \right]^T}
\end{equation}
where \({{\bf{W}}_1}^i\) is the weights of the \(i\)th neuron in the encoder layer, and \(b_1^{i}\) stands for the bias of the \(i\)th neurons in the encoder layer. Then, we should make these weights changing back to \(k_1 \times k_2\) dimensions to facilitate the next feature mapping operation.

The first de-noising auto encoders unsupervised layer's feature maps are generated by applying the encoder layer to convolute with the input images. For each input image, \(L_1\) feature maps can be obtained by
\begin{equation} \label{encodermapping}
\begin{array}{*{20}{c}}
   {{{\bf{I}}_1}^{(i)} = \sigma \left( {\bf{I}} * {{\bf{w}}_1}^i + {{b}}_1^i \bf{1} \right),} & {i = 1, \cdots ,{L_1}}  \\
\end{array},
\end{equation}
where \(\bf{I}\) stands for an input image with zero padded to make \({\bf{I}}_1^{(i)}\) have the same size as the input image, and \(\bf{1}\) is a matrix of the same size as \({\bf{I}}_1^{(i)}\) full of elements \(1\).

\textbf{The second unsupervised layer}

Patches should be extracted from feature maps \({\bf{I}}_1^{(i)}\) obtained from the first unsupervised layer. These patches are also the pre-processed by LCN and whitening operations. By flattening these patches, the patch matrix is constructed as
\begin{equation}
{{\bf{Z}}_2} = \left[ {\begin{array}{*{20}{c}}
   {{\textbf{z}_2}^{(1)},} & {{\textbf{z}_2}^{(2)},} & { \cdots ,} & {{\textbf{z}_2}^{(m)}}  \\
\end{array}} \right] \in {R^{{k_1}{k_2} \times m}},
\end{equation}
where \(\textbf{z}_2^{(i)}\) stands for flattened patch from the first layer's feature maps after pre-processing, and m stands for the number of extracted patches, and \(k_1\), \(k_2\) stand for the size of extracted patches.

The unsupervised learning method is applied to the patch matrix \({\bf{Z}}_2\) the same as the first layer. Assuming that the number of the second layer's filters is \(L_2\), its solution can be obtained as \ref{pcasolution} or \ref{encodersolution} by solving \ref{pcamodel} or \ref{encodermodel}. The second layer's feature maps based on the first layer's feature map are calculated as \ref{pcamapping} or \ref{encodermapping}.

If the used unsupervised learning is PCA, the second layer's unsupervised feature maps are calculated by
\begin{equation}
\begin{array}{*{20}{c}}
   {{{\bf{I}}_2}^{(i)} = {{\bf{I}}_1} * {{\bf{w}}_2}^i,} & {i = 1, \cdots ,{L_2}}  \\
\end{array},
\end{equation}
where \({\bf{I}}_1\) is the zero padded image of the first layer's feature map, and \({\bf{w}}_2^i\) is the second layer's feature bank weights.

If the unsupervised learning is auto encoder, the second layer's unsupervised feature maps are calculated by
\begin{equation} 
\begin{array}{*{20}{c}}
   {{{\bf{I}}_2}^{(i)} = \sigma \left( {{\bf{I}}_1} * {{\bf{w}}_2}^i + {{b}}_2^i \bf{1} \right),} & {i = 1, \cdots ,{L_2}}  \\
\end{array},
\end{equation}
where \({{\bf{w}}_2}^i\) and \({{b}}_2^i\) are the weights and bias of the \(i\)th neuron in the second auto encoder layer.

For an input image, we get \(L_1 \times (L_2+1) \) feature maps after the two convolution layers by concatenating the first layer maps to the second layer. That is


\begin{align}\label{1111111111}
  \{{{\bf{I}}_1}^{(1)},{{\bf{I}}_1}^{(2)}, \cdots   ,{{\bf{I}}_1}^{({L_1})},{{\bf{I}}_2}{{^{(1)}}_1},{{\bf{I}}_2}{{^{(2)}}_1}, \cdots, &\notag\\
  {{\bf{I}}_2}{{^{({L_2})}}_1}, \cdots ,{{\bf{I}}_2}{{^{(1)}}_{{L_1}}},
{{\bf{I}}_2}{{^{(2)}}_{{L_1}}}, \cdots ,{{\bf{I}}_2}{{^{({L_2})}}_{{L_1}}} & \},
\end{align}
where \({{\bf{I}}_1}^{(i)}\) stands for the \(i\)th feature map of the first layer, and \({{\bf{I}}_2}{^{(j)}}_i\) stands for the \(j\)th feature map of the second layer for the \(i\)th feature map of the first layer .

\subsubsection{Block-wise histogram}

\begin{figure}[!t]
\centering
\includegraphics{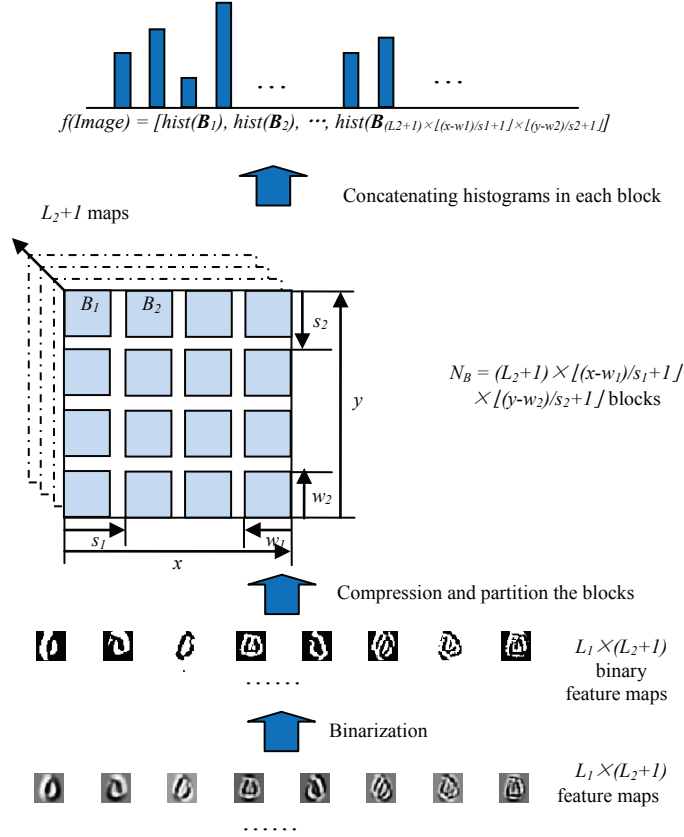}
\caption{Illustration showing the output layer of deep trans-layer unsupervised network. We first encoder the output of the trans-layer unsupervised learning. Then compress each \(L_1\) binarized feature maps into one feature map with pixels in an integer range from \(0\) to \(2^{L_1}-1\). Thus, \(L_2+1\) compressed feature maps will be obtained for each image. Then use a \(w_1-by-w_2\) receptive field and stride \(s\) to generate \( \lfloor (x-w_1)/s_1+1 \rfloor \times \lfloor (y-w_2)/s_2+1 \rfloor \) blocks for each compressed feature map. For each block, histogram with \(2^{L_1}\) bins is calculated. Finally, concatenate the \(N_B\) histograms to form the feature representation of deep trans-layer unsupervised network.}
\label{DeepPCAOut}
\end{figure}

The third layer of the network is illustrated in Figure \ref{DeepPCAOut}. The third layer is quite similar to that of LBPH \cite{LBPH}. For an input training image, the first step is to encoder the \(L_1 \times (L_2+1) \) real valued feature maps with binary values, 0 and 1. The operation converts these feature maps into binary images.

The second step is to compress these binary feature maps by quantizing each \(L_1\) binary feature maps. Here the number of second layer filters \(L_2\) is set as 8, and the number of first layer filters \(L_1\) is also set as 8. That is, we compress each \(L_1\) binary feature maps into one feature map, and the compressed feature maps have pixel values from 0 to 255. Then we get \(L_2+1\) compressed feature maps for each training image as
\begin{equation}
\left\{ {{{\bf{I}}_c}^{(1)},{{\bf{I}}_c}^{(2)}, \cdots ,{{\bf{I}}_c}^{({L_2})},{{\bf{I}}_c}^{({L_2} + 1)}} \right\},
\end{equation}
where \({I_c}^{(i)}\) represents the \(i\)th compressed feature map.

The third step is to construct block-wise histogram illustrated as the third procedure in Figure \ref{DeepPCAOut}. First, we should partition each compressed feature map. Assuming that the size of compressed feature is \(x \times y\), and the size of block is \(w_1 \times w_2\) with strides \(s_1 \times s_2\), each compressed feature map is partitioned into \(\lfloor (x-w_1)/s_1+1 \rfloor \times \lfloor (y-w_2)/s_2+1 \rfloor\) blocks. For all the \(L_2+1\) compressed feature maps of each input image, we get blocks as
\begin{equation}
\left\{ {{{\bf{B}}_1},{{\bf{B}}_2}, \cdots ,{{\bf{B}}_{({L_2} + 1) \times \lfloor (x - {w_1})/{s_1} + 1 \rfloor \times \lfloor (y - {w_2})/{s_2} + 1\rfloor}}} \right\},
\end{equation}
where \({\bf{B}}_i\) stands for the \(i\)th block. Next step is to build histograms in each of the blocks. In the deep trans-layer PCA network, we set the number of bins in the histogram to \(2^{L_1}\). It means that each integer of the pixel values is set as a bin and a sparse vector representing the histogram is constructed. Then concatenate these \(N_B = ({L_2} + 1) \times \lfloor (x - {w_1})/{s_1} + 1 \rfloor \times \lfloor (y - {w_2})/{s_2} + 1\) histograms to form a more complete representation of the input image as
\begin{equation}
\begin{aligned}
&f(Image) = \\
&{[hist{({{\bf{B}}_1})^T},hist{({{\bf{B}}_2})^T}, \cdots ,hist{({{\bf{B}}_{{N_B}}})^T}]^T} \in {R^{{N_B}({2^{{L_1}}})}},
\end{aligned}
\end{equation}
where \(hist()\) stands for histogram operators. Then we use the deep trans-layer unsupervised network representation of each training image to learn a dimensional reduction weight, or to train a classifier to tackle the next applications directly.

\subsection{Feature extraction and classification}

After the above representation learning phase, the unsupervised convolution feature banks have been learned. Given a test image, the feature extraction is to map the image to trans-layer representation with \({N_B}({2^{{L_1}}})\) dimensions by these feature banks.

\subsubsection{Convolutional extraction and block-wise histogram}

Given a test image, we should zero-pad the image for the sake of keeping the same size between feature maps and input image. Then the pre-processing of LCN and whitening is applied to the zero-padded image. Feature maps are calculated by the first unsupervised layer with \(L_1\) filters of size \(k_1 \times k_2\).

For the second unsupervised layer, the procedure is the same as the first layer. First, zero-pad each feature map. Then, pre-process these feature maps. Next, put these feature maps into unsupervised learning of \(L_2\) feature banks of size \(k_1 \times k_2\). Finally, merge the first layer's feature maps with the second layer's feature maps to form a more complete trans-layer representation.

The last phase is block-wise histogram. Binary encoding is used to tackle these real valued feature maps. And then form a compact representation by quantizing each \(L_1\) binary feature maps into one feature map with pixel values from 0 to \(2^{L_1}-1\). Then partition these compact feature maps into blocks, and construct histogram representation within each block. Concatenate these histograms to form the deep trans-layer unsupervised representation of translation and rotation invariance.

\subsubsection{Classification}

For real applications in the next experiments, classifiers or dimension reduction methods are following the deep trans-layer unsupervised representation. Due to the relative high dimensions of this representation, whitening PCA (WPCA) is used to reduce the representation in object recognition tasks. The WPCA is conducted by conventional PCA weights weighted by the inverse of their corresponding squared root energies. Also, the deep trans-layer unsupervised representation can be directly used to train a classifier to tackle recognition tasks. In our experiments of digit recognition and object recognition, a simple linear SVM classifier with no parameter tuned is used following the deep trans-layer unsupervised network. The parameter, the cost factor \(C\), in the used linear SVM software kit LIBLINEAR is 1 as default \cite{LibLinear}.

\section{Experimental results}\label{experiment}

In the experiment, we will validate the performance of deep trans-layer PCA network (using PCA unsupervised learning the local receptive features) and deep trans-layer auto encoder network (using de-noising auto encoder unsupervised learning the local receptive features). The proposed deep trans-layer unsupervised network has two key phases, LCN in pre-processing and trans-layer concatenation. We will validate the two phases in the \emph{MNIST} variations data set \cite{Bengio_ICML07} using the deep tran-layer PCA network. Also, other parameters such as block size and stride size are chosen through cross validation or validation set. Benchmark experiments are conducted on digit recognition of \emph{MNIST} \cite{LeCun_98} and \emph{MNIST} variations \cite{Bengio_ICML07} data sets, and object recognition of Caltech 101 data set \cite{LiFeifei_CVPR04}.

\subsection{Effect of local contrast normalization}\label{sec:expLCN}

We first validate the effect of LCN followed by conventional PCA in our network. Experiments about deep trans-layer PCA network with LCN and without LCN are conducted on the \emph{MNIST} variations data sets.

The \emph{MNIST} data set contains 60,000 training samples and 10,000 test samples of gray images with size \(28 \times 28\) pixels. The data set is a subset of \emph{NIST}, which contains hand-written digits in real world. The recognition targets have been size-normalized and centered in the images \cite{LeCun_98}. \emph{MNIST} variations data sets are created by applying simple controllable factor variations on \emph{MNIST} digits \cite{Bengio_ICML07}. The data sets are effective ways to study the invariance ability of representation learning methods. Details about recognition tasks, numbers of classes and samples are included in Table \ref{tab:dataset}.

%
\begin{table}[!t]
\caption{Details of the 9 used data sets for digit recognition on \emph{MNIST} \cite{LeCun_98} and \emph{MNIST} variations \cite{Bengio_ICML07}}
\label{tab:dataset}       
\centering
\begin{tabular}{|c|c|c|c|}
\hline
Data sets & Recognition tasks & \#Classes & \#Train-Test \\
\hline
\emph{MNIST} & Handwritten digits from 0 to 9 & 10 & 60000-10000 \\
\hline
\emph{mnist-basic} & Smaller subset of \emph{MNIST} & 10 & 12000-50000 \\
\hline
\emph{mnist-rot} & Same as basic with rotation & 10 & 12000-50000 \\
\hline
\emph{mnist-back-rand} & Same as basic with random background & 10 & 12000-50000 \\
\hline
\emph{mnist-back-image} & Same as basic with image background & 10 & 12000-50000 \\
\hline
\emph{mnist-rot-back-image} & Same as basic with rotation and image background & 10 & 12000-50000 \\
\hline
\emph{rectangles} & Tall or wide rectangles & 2 & 1200-50000 \\
\hline
\emph{rectangles-image} & Same as rect. with image background & 2 & 12000-50000 \\
\hline
\emph{ConvexNonConvex} & Convex or non-convex shapes & 2 & 8000-50000 \\
\hline
\end{tabular}
\end{table}

The parameters of the block size, the stride size and the patch size are determined on validation set experimentally. The validation sets are typically partitioned from the training set in consistence with the related work \cite{Bengio_ICML07}. On the digit recognition tasks, the filter size is set to \(7 \times 7\) pixels, the number of filters is set to \(L_1\)=\(L_2\)=8 and the size of strides is the half size of the block. For \emph{MNIST}, \emph{MNIST basic}, \emph{mnist-rotation}, and \emph{rectangles-image} data sets, the block size is \(7 \times 7\) pixels. For \emph{mnist-back-rand}, \emph{mnist-back-image} and \emph{mnist-rot-back-image} data sets, the block size is \(4 \times 4\) pixels. For \emph{rectangles} data set, the block size is \(14 \times 14\) pixels. For \emph{convex} data set, the block size is \(28 \times 28\) pixels. These validated tiny parameters are fixed in the following digit recognition tasks. A simple linear SVM with default parameters is connected to the deep trans-layer PCA representation to do recognition task \cite{LibLinear}.

Two groups of experiments are conducted on \emph{MNIST basic}, \emph{mnist-rotation}, \emph{mnist-back-rand}, \emph{mnist-back-image-rotation} data sets to validate the effect of LCN. The performance of deep trans-layer PCA Network with and without LCN is reported in Table \ref{tab:LCN}.
%
\begin{table}[!t]
\caption{Comparison of classification error rates (\%) on \emph{MNIST} variations data sets using deep trans-layer PCA network with LCN and without LCN.}
\label{tab:LCN}       
\centering
\begin{tabular}{|c|c|c|c|c|c|}
\hline
Methods & \emph{Basic} & \emph{Rot.} & \emph{Bk-rand} & \emph{Bk-im} & \emph{Bk-im-rot} \\
\hline
\emph{Without LCN} & 1.03 & 6.40 & \textbf{5.09} & 10.51 & 34.79 \\
\hline
\emph{With LCN} & \textbf{0.98} & \textbf{6.36} & 5.22 & \textbf{9.95} & \textbf{33.55} \\
\hline
\end{tabular}
\end{table}

From Table \ref{tab:LCN}, we observe that our model with LCN achieves better performance on these data sets except for \emph{mnist-back-rand} data set. The results prove that LCN helps conventional PCA performance, which is explained in section \ref{sec:preprocess}. The main reason why LCN performs worse in \emph{mnist-back-rand} data set is that, the mean and standard deviation in the data set are all corrupted noise information due to large area random noise in the background, which degrades the performance. The LCN boosts the performance in natural images, which is validated by the improvements in \emph{MNIST basic}, \emph{mnist-rotation}, \emph{mnist-back-image} and \emph{mnist-back-image-rotation} data sets.

\subsection{Effect of trans-layer connection}

The effect of trans-layer connection is validated on \emph{MNIST basic}, \emph{mnist-rotation}, \emph{mnist-back-rand}, \emph{mnist-back-image-rotation} data sets in the second experiment. The parameters are set as that of section \ref{sec:expLCN}. The performance of deep trans-layer PCA network with and without trans-layer connection is recorded in Table \ref{tab:trans}.
%
\begin{table}[!t]
\caption{Comparison of classification error rates (\%) on \emph{MNIST} variations data sets using deep trans-layer PCA network with and without trans-layer connection.}
\label{tab:trans}       
\centering
\begin{tabular}{|c|c|c|c|c|c|}
\hline
Methods & \emph{Basic} & \emph{Rot.} & \emph{Bk-rand} & \emph{Bk-im} & \emph{Bk-im-rot} \\
\hline
\emph{Without trans-layer connection} & 1.05 & 6.62 & 5.94 & 10.58 & 34.58 \\
\hline
\emph{With trans-layer connection} & \textbf{0.98} & \textbf{6.36} & \textbf{5.22} & \textbf{9.95} & \textbf{33.55} \\
\hline
\end{tabular}
\end{table}

From Table \ref{tab:trans}, we observe that deep trans-layer PCA networks with trans-layer connection have a consistently better performance than those of without trans-layer connection. Trans-layer connection boosts almost 0.1 percentage performance even on the hardest task of \emph{MNIST basic} data set, and 1\% performance for \emph{mnist-back-image-rotation} data set. The trans-layer connection in deep trans-layer PCA network provides a more complete representation that is always helpful for recognition.

\subsection{Digit recognition on \emph{MNIST} and \emph{MNIST} variations data sets}

We report the performance of the implemented model on \emph{MNIST} and \emph{MNIST} variations data sets compared with other methods such as convolution network (ConvNet) \cite{LeCun_ICCV09} and ScatNet-2 \cite{Mallat_PAMI13}. The performance of these methods is recorded in Table \ref{tab:mnist} of \emph{MNIST} data set and Table \ref{tab:mnistvar} of \emph{MNIST} variations data sets respectively. For fair comparison, the following results do not include the results using augmented samples.
%
\begin{table}[!t]
\caption{\emph{MNIST} classification error rates (\%). Note that the compared methods are not including methods that augment the training data.}
\label{tab:mnist}       
\centering
\begin{tabular}{|c|c|}
\hline
Methods & \emph{MNIST} error rates (\%) \\
\hline
K-NN-SCM \cite{Shape_PAMI02} & 0.63 \\
\hline
K-NN-IDM \cite{Deform_PAMI07} & 0.54 \\
\hline
CDBN \cite{Andrew_ICML09} & 0.82 \\
\hline
HSC \cite{Yukai_CVPR11} & 0.77 \\
\hline
ConvNet \cite{LeCun_ICCV09} & 0.53 \\
\hline
Stochastic Pooling ConvNet \cite{Fergus_ICLR13} & 0.47 \\
\hline
Conv. Maxout + Dropout \cite{Bengio_ICML13} & 0.45 \\
\hline
ScatNet-2 (\(SVM_{rbf}\)) \cite{Mallat_PAMI13} & \textbf{0.43} \\
\hline
PCANet \cite{Mayi_PCANet} & 0.66 \\
\hline
Deep Trans-layer PCA Network & 0.58 \footnotemark[1] \\
\hline
Deep Trans-layer Autoencoder Network & 0.55 \footnotemark[1] \\
\hline
\end{tabular}
\end{table}
\setcounter{footnote}{0}
\footnotetext[1]{The matlab code can be downloaded from \url{https://github.com/wentaozhu/Deep-trans-layer-unsupervised-network.git}}
%
\begin{table}[!t]
\caption{Comparison of classification error rates (\%) on \emph{MNIST} variations data sets.}
\label{tab:mnistvar}       
\centering
\begin{tabular}{|c|c|c|c|c|c|c|c|c|}
\hline
Methods & \emph{Basic} & \emph{Rot.} & \emph{Bk-rand} & \emph{Bk-im} & \emph{Bk-im-rot} & \emph{Rect} & \emph{Rect-im} & \emph{Con.}  \\
\hline
CAE-2 \cite{Bengio_ICML11} & 2.48 & 9.66 & 10.90 & 15.50 & 45.23 & 1.21 & 21.54 & - \\
\hline
TIRBM \cite{Lee_ICML12} & - & \textbf{4.20} & - & - & 35.50 & - & - & - \\
\hline
\tabincell{c}{PGBM\\+DN-1 \cite{Lee_ICML13}} & - & - & 6.08 & 12.25 & 36.76 & - & - & - \\
\hline
\tabincell{c}{ScatNet-2\\(\(SVM_{rbf}\)) \cite{Mallat_PAMI13}} & 1.27 & 7.48 & 18.40 & 12.30 & 50.48 & \textbf{0.01} & \textbf{8.02} & 6.50 \\
\hline
PCANet \cite{Mayi_PCANet} & 1.06 & 7.37 & 6.19 & 10.95 & 35.48 & 0.24 & 14.00 & 4.36\\
\hline
\tabincell{c}{Deep Trans-layer\\ PCA Network} & \textbf{0.98} & 6.36 & \textbf{5.22} & \textbf{9.95} & \textbf{33.55} & 0.09 & 13.18 & \textbf{3.52} \\
\hline
\tabincell{c}{Deep Trans-layer \\ Autoencoder Network} & \textbf{1.02} & 7.56 & \textbf{5.56} & \textbf{8.37} & \textbf{34.43} & \textbf{0.01} & 13.01 & \textbf{2.90} \\
\hline
\end{tabular}
\end{table}

From Table \ref{tab:mnist}, the results show that the deep trans-layer unsupervised network is only inferior to ScatNet-2 and enhanced Convolution Network related methods. It is worthy to mention that the performance of ScatNet-2 is achieved by connected with a non-linear SVM with RBF kernels with tuned parameters, but our model is connected with a linear SVM with all default parameters in LIBLINEAR software kit \cite{LibLinear}. Our model's performance (0.55) is highly comparable to that of Convolution Network (0.53) on \emph{MNIST} data set \cite{LeCun_ICCV09}. Because \emph{MNIST} data set contains too many training samples with small intra-class variability, most of methods work well on this data set and the tiny difference is not much meaningful statistically. Despite this, the deep trans-layer PCA network boosts almost 0.1\% performance higher than that of PCA related method, PCANet.

From Table \ref{tab:mnistvar}, we observe that the deep trans-layer unsupervised network achieves the best performance on six data sets with a simple linear SVM classifier. Our model has a highly superior performance than other methods. It is sufficient to prove that the proposed structure of LCN pre-processing and trans-layer connection work well in the convolution structure with unsupervised filters.

\subsection{Object recognition on Caltech 101 data set}

We also evaluate the model for object recognition task on Caltech 101 data set. Caltech 101 data set contains color images belonging to 102 categories including a background class. The number of each class's images varies from 31 to 800 \cite{LiFeifei_CVPR04}. The pre-processing of the data set is to convert the images into gray scale, and adjust the longer side of the image to 300 with preserved aspect ratio. Two typical tasks are conducted. One is with a training set of 15 samples per class. The other is with training set of 30 samples per class. The training sets are randomly sampled from Caltech 101, and the rest are test set. Five rounds of experiments are recorded, and the performance is recorded as the average of the five rounds of results.

Parameters are set as follows. The filter size is set to \(7 \times 7\) pixels, the number of filters is set to \(L_1 = L_2 = 8\), the block size is set to a quarter of the image size and the size of strides is set to the half size of the blocks. The WPCA is used to reduce the dimension of each block's trans-layer representation from 256 to 64. Linear SVM with default parameters in LIBLINEAR \cite{LibLinear} is used to tackle the recognition task. Comparison results on gray raw images are recorded in Table \ref{tab:caltech}.
%
\begin{table}[!t]
\caption{Comparison of classification accuracy (\%) in terms of mean and stand deviation based on gray-level images of Caltech 101 data set.}
\label{tab:caltech}       
\centering
\begin{tabular}{|c|c|c|}
\hline
Methods & 15 samples per class (\%) & 30 samples per class (\%) \\
\hline
CDBN \cite{Andrew_ICML09} & 57.70 \(\pm\) 1.50 & 65.40 \(\pm\) 0.50 \\
\hline
ConvNet \cite{LeCun_NIPS10} & 57.60 \(\pm\) 0.40 & 66.30 \(\pm\) 1.50 \\
\hline
DeconvNet \cite{Fergus_CVPR10} & 58.60 \(\pm\) 0.70 & 66.90 \(\pm\) 1.10 \\
\hline
Chen et al. \cite{Chen_ICML11} & 58.20 \(\pm\) 1.20 & 65.80 \(\pm\) 0.60 \\
\hline
Zou et al. \cite{Andrew_NIPS12} & - & 66.50 \\
\hline
HSC \cite{Yukai_CVPR11} & - & 74.0 \\
\hline
PCANet \cite{Mayi_PCANet} & 61.46 \(\pm\) 0.76 & 68.56 \(\pm\) 1.01 \\
\hline
\tabincell{c}{Deep Trans-layer\\ PCA Network} & \textbf{67.11 \(\pm\) 0.64} & \textbf{75.98 \(\pm\) 0.58} \\
\hline
\tabincell{c}{Deep Trans-layer \\ Autoencoder Network} & \textbf{67.03 \(\pm\) 0.56} & \textbf{75.90 \(\pm\) 0.19} \\
\hline
\end{tabular}
\end{table}

The Table \ref{tab:caltech} shows that the deep trans-layer unsupervised network gets the impressive performance trained by 15 samples per class and 30 samples per class tasks respectively by simply using the unsupervised methods. More surprisingly, our model with no elaborately tuned parameters gets about 2\% upper than that of HSC \cite{Yukai_CVPR11} on 30 samples per class task. The proposed simple network really makes a high progress for the data set.

\subsection{Face verification on LFW-a data set}

Face verification task is conducted with our model on LFW-a data set \cite{Wolf_PAMI11}. The LFW data set contains more than 13,000 faces of 5,749 different individuals, and these images were collected from the web of unconstrained conditions. In the LFW data set, the unsupervised setting is used for the deep trans-layer PCA network for sufficiently validating of representation effectiveness. The LFW-a data set with alignment is used, and we cropped the face images into \(150 \times 80\) pixels. The standard evaluation protocol on LFW is followed for performance evaluation. The histogram block size is \(15 \times 13\) with non-overlapping. Other parameters are set the same as before. The WPCA is used to reduce the dimension of trans-layer PCA representation to 3,200 after additional square-root operation in the data set. Then use NN classifier with cosine distance to tackle the verification task. The performance is recorded in Table 7 for \emph{unsupervised setting with a single descriptor}.
%
\begin{table}[!t]
\caption{Comparison of verification rates (\%) on LFW-a data set under unsupervised setting with a single descriptor.}
\label{tab:7}       
\centering
\begin{tabular}{|c|c|}
\hline
Methods & Verification accuracy (\%) \\
\hline
POEM \cite{Caplier_TIP12} & 82.70 \(\pm\) 0.59 \\
\hline
High-dim. LBP \cite{Sunjian_CVPR13} & 84.08 \\
\hline
High-dim. LE \cite{Sunjian_CVPR13} & 84.58 \\
\hline
SFRD \cite{Cuizhen_CVPR13} & 84.81 \\
\hline
I-LQP \cite{Jurie_BMVC12} & 86.20 \(\pm\) 0.46 \\
\hline
OCLBP \cite{Wolf_ICCV13} & 86.66 \(\pm\) 0.30 \\
\hline
PCANet \cite{Mayi_PCANet} & 86.28 \(\pm\) 1.14 \\
\hline
Deep Trans-layer PCA Network & \textbf{87.10 \(\pm\) 0.43} \\
\hline
\end{tabular}
\end{table}

Table 7 shows that the deep trans-layer PCA network achieves the best performance with the unsupervised setting on the LFW-a data set. Also, the boosted performance of our model in contrast to PCANet with the same 3,200 dimensions reveals that, the trans-layer architecture provides uncorrelated information in the increased dimensions, which is beneficial for tasks.

\section{Discussion}\label{discus}
One of the typical cases to illustrate why the trans-layer representation works is that the local information of naevi in our faces may vanish in the upper layer's unsupervised representation for the size of receptive fields are increased and the local discriminative information may be neglected. However, the naevi is of highly discriminative features to recognize each person. In conventional deep neural networks, such discriminative details also hard to preserve with the increasing of layer numbers. The trans-layer representation from the bottom layer preserves these details and is helpful for recognition.

\begin{figure}[!t]
\centering
\includegraphics[width=3.7in]{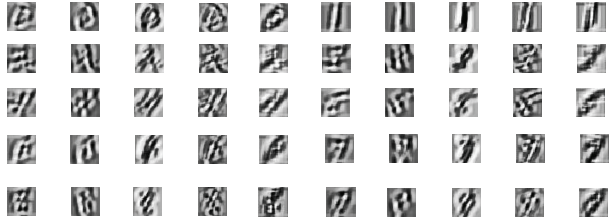}
\caption{Illustration of the second layer's feature maps on \emph{MNIST} data set, five maps per category. We can hardly recognize most maps for much information loss in the multi-layer PCA network.}
\label{SecondLayerMNIST}
\end{figure}

The trans-layer representation scheme is quite successful for the convolution network of PCA and auto encoder filters, because the number of filters is relatively too small to preserve enough information for tasks. And information will be lost rapidly with the increase of layer numbers, as illustrated in Fig. \ref{SecondLayerMNIST}. If the model uses only the second layer's feature maps, there would be much noise and lose much useful information. Therefore, the second layer's feature maps are not enough for classification, and the trans-layer connection is quite helpful.

The deep trans-layer PCA network also has some problems to be improved despite that LCN and trans-layer representation are added into it. The main problem is the high dimensions of the trans-layer representation. Although good representation for tasks needs high dimensions, we should make the dimensions of representation as low as possible providing good representational ability. So the future work could include that other translation and rotation invariance preserving methods to reduce the dimensions of the trans-layer representations such as pooling operations \cite{LeCun_ICCV09, LeCun_98, Fergus_ICLR13}.

\section{Conclusion}\label{con}
In this paper, a novel feature learning and representation model, the deep trans-layer unsupervised network, is demonstrated. Several key elements are added to boost the deep unsupervised learning such as the LCN operation and trans-layer representation. Several experiments on digit recognition and object recognition tasks have shown that, the proposed method based on layer-wise unsupervised learning the local receptive features also obtains the impressive results by trans-layer representation. The deep trans-layer unsupervised network has a quite simple structure with much few parameters, in which only the cascaded local receptive filters need to be learned by unsupervised methods. The structure of trans-layer network also provides a novel direction for us to mine.


%

%


%

\section*{References}

\bibliography{mybibfile}

\begin{thebibliography}{10}
\expandafter\ifx\csname url\endcsname\relax
  \def\url#1{\texttt{#1}}\fi
\expandafter\ifx\csname urlprefix\endcsname\relax\def\urlprefix{URL }\fi
\expandafter\ifx\csname href\endcsname\relax
  \def\href#1#2{#2} \def\path#1{#1}\fi

\bibitem{Lowe_IJCV04}
D.~G. Lowe, Distinctive image features from scale-invariant keypoints,
  International journal of computer vision 60~(2) (2004) 91--110.

\bibitem{Triggs_CVPR05}
N.~Dalal, B.~Triggs, Histograms of oriented gradients for human detection, in:
  Computer Vision and Pattern Recognition, 2005. CVPR 2005. IEEE Computer
  Society Conference on, Vol.~1, IEEE, 2005, pp. 886--893.

\bibitem{Hinton_SCI06}
G.~E. Hinton, R.~R. Salakhutdinov, Reducing the dimensionality of data with
  neural networks, Science 313~(5786) (2006) 504--507.

\bibitem{Mayi_PCANet}
T.-H. Chan, K.~Jia, S.~Gao, J.~Lu, Z.~Zeng, Y.~Ma, Pcanet: A simple deep
  learning baseline for image classification?, arXiv preprint arXiv:1404.3606.

\bibitem{LeCun_ICCV09}
K.~Jarrett, K.~Kavukcuoglu, M.~Ranzato, Y.~LeCun, What is the best multi-stage
  architecture for object recognition?, in: Computer Vision, 2009 IEEE 12th
  International Conference on, IEEE, 2009, pp. 2146--2153.

\bibitem{Coates_KMEANSJ}
A.~Coates, A.~Y. Ng, Learning feature representations with k-means, in: Neural
  Networks: Tricks of the Trade, Springer, 2012, pp. 561--580.

\bibitem{LiFeifei_CVPR04}
L.~Fei-Fei, R.~Fergus, P.~Perona, Learning generative visual models from few
  training examples: An incremental bayesian approach tested on 101 object
  categories, Computer Vision and Image Understanding 106~(1) (2007) 59--70.

\bibitem{LBPH}
T.~Ojala, M.~Pietikainen, T.~Maenpaa, Multiresolution gray-scale and rotation
  invariant texture classification with local binary patterns, Pattern Analysis
  and Machine Intelligence, IEEE Transactions on 24~(7) (2002) 971--987.

\bibitem{Hinton_NIPS12}
A.~Krizhevsky, I.~Sutskever, G.~E. Hinton, Imagenet classification with deep
  convolutional neural networks, in: Advances in neural information processing
  systems, 2012, pp. 1097--1105.

\bibitem{LeCun_ICML13}
L.~Wan, M.~Zeiler, S.~Zhang, Y.~L. Cun, R.~Fergus, Regularization of neural
  networks using dropconnect, in: Proceedings of the 30th International
  Conference on Machine Learning (ICML-13), 2013, pp. 1058--1066.

\bibitem{Andrew_NIPS11}
Q.~V. Le, A.~Karpenko, J.~Ngiam, A.~Y. Ng, Ica with reconstruction cost for
  efficient overcomplete feature learning, in: Advances in Neural Information
  Processing Systems, 2011, pp. 1017--1025.

\bibitem{Mallat_PAMI13}
J.~Bruna, S.~Mallat, Invariant scattering convolution networks, Pattern
  Analysis and Machine Intelligence, IEEE Transactions on 35~(8) (2013)
  1872--1886.

\bibitem{Yukai_CVPR11}
K.~Yu, Y.~Lin, J.~Lafferty, Learning image representations from the pixel level
  via hierarchical sparse coding, in: Computer Vision and Pattern Recognition
  (CVPR), 2011 IEEE Conference on, IEEE, 2011, pp. 1713--1720.

\bibitem{LeCun_CVPR10}
Y.-L. Boureau, F.~Bach, Y.~LeCun, J.~Ponce, Learning mid-level features for
  recognition, in: Computer Vision and Pattern Recognition (CVPR), 2010 IEEE
  Conference on, IEEE, 2010, pp. 2559--2566.

\bibitem{Andrew_ICML09}
H.~Lee, R.~Grosse, R.~Ranganath, A.~Y. Ng, Convolutional deep belief networks
  for scalable unsupervised learning of hierarchical representations, in:
  Proceedings of the 26th Annual International Conference on Machine Learning,
  ACM, 2009, pp. 609--616.

\bibitem{LibLinear}
R.-E. Fan, K.-W. Chang, C.-J. Hsieh, X.-R. Wang, C.-J. Lin, Liblinear: A
  library for large linear classification, The Journal of Machine Learning
  Research 9 (2008) 1871--1874.

\bibitem{Bengio_ICML07}
H.~Larochelle, D.~Erhan, A.~Courville, J.~Bergstra, Y.~Bengio, An empirical
  evaluation of deep architectures on problems with many factors of variation,
  in: Proceedings of the 24th international conference on Machine learning,
  ACM, 2007, pp. 473--480.

\bibitem{LeCun_98}
Y.~LeCun, L.~Bottou, Y.~Bengio, P.~Haffner, Gradient-based learning applied to
  document recognition, Proceedings of the IEEE 86~(11) (1998) 2278--2324.

\bibitem{Shape_PAMI02}
S.~Belongie, J.~Malik, J.~Puzicha, Shape matching and object recognition using
  shape contexts, Pattern Analysis and Machine Intelligence, IEEE Transactions
  on 24~(4) (2002) 509--522.

\bibitem{Deform_PAMI07}
D.~Keysers, T.~Deselaers, C.~Gollan, H.~Ney, Deformation models for image
  recognition, Pattern Analysis and Machine Intelligence, IEEE Transactions on
  29~(8) (2007) 1422--1435.

\bibitem{Fergus_ICLR13}
M.~D. Zeiler, R.~Fergus, Stochastic pooling for regularization of deep
  convolutional neural networks, arXiv preprint arXiv:1301.3557.

\bibitem{Bengio_ICML13}
I.~J. Goodfellow, D.~Warde-Farley, M.~Mirza, A.~Courville, Y.~Bengio, Maxout
  networks, arXiv preprint arXiv:1302.4389.

\bibitem{Bengio_ICML11}
S.~Rifai, P.~Vincent, X.~Muller, X.~Glorot, Y.~Bengio, Contractive
  auto-encoders: Explicit invariance during feature extraction, in: Proceedings
  of the 28th International Conference on Machine Learning (ICML-11), 2011, pp.
  833--840.

\bibitem{Lee_ICML12}
K.~Sohn, H.~Lee, Learning invariant representations with local transformations,
  arXiv preprint arXiv:1206.6418.

\bibitem{Lee_ICML13}
K.~Sohn, G.~Zhou, C.~Lee, H.~Lee, Learning and selecting features jointly with
  point-wise gated $\{$B$\}$ oltzmann machines, in: Proceedings of The 30th
  International Conference on Machine Learning, 2013, pp. 217--225.

\bibitem{LeCun_NIPS10}
K.~Kavukcuoglu, P.~Sermanet, Y.-L. Boureau, K.~Gregor, M.~Mathieu, Y.~L. Cun,
  Learning convolutional feature hierarchies for visual recognition, in:
  Advances in neural information processing systems, 2010, pp. 1090--1098.

\bibitem{Fergus_CVPR10}
M.~D. Zeiler, D.~Krishnan, G.~W. Taylor, R.~Fergus, Deconvolutional networks,
  in: Computer Vision and Pattern Recognition (CVPR), 2010 IEEE Conference on,
  IEEE, 2010, pp. 2528--2535.

\bibitem{Chen_ICML11}
B.~Chen, G.~Polatkan, G.~Sapiro, L.~Carin, D.~B. Dunson, The hierarchical beta
  process for convolutional factor analysis and deep learning, in: Proceedings
  of the 28th International Conference on Machine Learning (ICML-11), 2011, pp.
  361--368.

\bibitem{Andrew_NIPS12}
W.~Zou, S.~Zhu, K.~Yu, A.~Y. Ng, Deep learning of invariant features via
  simulated fixations in video, in: Advances in Neural Information Processing
  Systems, 2012, pp. 3212--3220.

\bibitem{Wolf_PAMI11}
L.~Wolf, T.~Hassner, Y.~Taigman, Effective unconstrained face recognition by
  combining multiple descriptors and learned background statistics, Pattern
  Analysis and Machine Intelligence, IEEE Transactions on 33~(10) (2011)
  1978--1990.

\bibitem{Caplier_TIP12}
N.-S. Vu, A.~Caplier, Enhanced patterns of oriented edge magnitudes for face
  recognition and image matching, Image Processing, IEEE Transactions on 21~(3)
  (2012) 1352--1365.

\bibitem{Sunjian_CVPR13}
D.~Chen, X.~Cao, F.~Wen, J.~Sun, Blessing of dimensionality: High-dimensional
  feature and its efficient compression for face verification, in: Computer
  Vision and Pattern Recognition (CVPR), 2013 IEEE Conference on, IEEE, 2013,
  pp. 3025--3032.

\bibitem{Cuizhen_CVPR13}
Z.~Cui, W.~Li, D.~Xu, S.~Shan, X.~Chen, Fusing robust face region descriptors
  via multiple metric learning for face recognition in the wild, in: Computer
  Vision and Pattern Recognition (CVPR), 2013 IEEE Conference on, IEEE, 2013,
  pp. 3554--3561.

\bibitem{Jurie_BMVC12}
S.~U. Hussain, T.~Napol{\'e}on, F.~Jurie, et~al., Face recognition using local
  quantized patterns, in: British Machive Vision Conference, 2012.

\bibitem{Wolf_ICCV13}
O.~Barkan, J.~Weill, L.~Wolf, H.~Aronowitz, Fast high dimensional vector
  multiplication face recognition, in: Computer Vision (ICCV), 2013 IEEE
  International Conference on, IEEE, 2013, pp. 1960--1967.

\end{thebibliography}

\end{document}